\documentclass[11pt,a4paper]{article}
\usepackage[hyperref]{emnlp2020}
\usepackage{times}
\usepackage{latexsym}

\usepackage{microtype}

\usepackage{graphicx}
\usepackage{float}
\usepackage{url}
\usepackage{booktabs}
\usepackage{hyperref}
\usepackage[utf8]{inputenc}
\usepackage{amsmath}
\usepackage{amsthm}
\usepackage{booktabs}
\usepackage{algorithm}
\usepackage{algorithmic}
\usepackage{verbatim}
\usepackage{textcomp}
\usepackage{amssymb}
\usepackage{pifont}
\newcommand{\cmark}{\ding{51}}
\newcommand{\xmark}{\ding{55}}
\newcommand\tab[1][2.5cm]{\hspace*{#1}}
\usepackage{lscape} 
\usepackage{listings}
\usepackage{xcolor}
\lstdefinelanguage{json}{
    basicstyle=\normalfont\ttfamily,
    numbers=left,
    numberstyle=\scriptsize,
    stepnumber=1,
    numbersep=8pt,
    showstringspaces=false,
    breaklines=false,
    frame=lines,
}

\aclfinalcopy 
\def\aclpaperid{3392} 

\title{Visuo-Linguistic Question Answering (VLQA) Challenge}

\author{Shailaja Keyur Sampat, Yezhou Yang and Chitta Baral \\
  Arizona State University \\
  Tempe, AZ, USA \\
  \texttt{\{ssampa17\thanks{corresponding author}, yz.yang, chitta\}@asu.edu} 
 }

\date{}

\begin{document}
\maketitle
\begin{abstract}
Understanding images and text together is an important aspect of cognition and building advanced Artificial Intelligence (AI) systems. As a community, we have achieved good benchmarks over language and vision domains separately, however joint reasoning is still a challenge for state-of-the-art computer vision and natural language processing (NLP) systems. 
We propose a novel task to derive joint inference about a given image-text modality and compile the Visuo-Linguistic Question Answering (VLQA) challenge corpus in a question answering setting. Each dataset item consists of an image and a reading passage, where questions are designed to combine both visual and textual information i.e., ignoring either modality would make the question unanswerable. We first explore the best existing vision-language architectures to solve VLQA subsets and show that they are unable to reason well. We then develop a modular method with slightly better baseline performance, but it is still far behind human performance. We believe that VLQA will be a good benchmark for reasoning over a visuo-linguistic context. The dataset, code and leaderboard is available at \url{https://shailaja183.github.io/vlqa/}.
\end{abstract}

\section{Introduction}
\begin{figure}[h]
  \center
  \includegraphics[width=\linewidth]{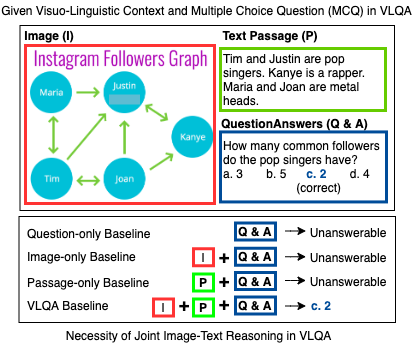}
  \caption{Example of Visuo-Linguistic Question Answering (VLQA) task for joint reasoning over image-text context.}
  \label{fig:coverexample}
\end{figure}

\begin{figure*}
  \center
  \includegraphics[width=\linewidth]{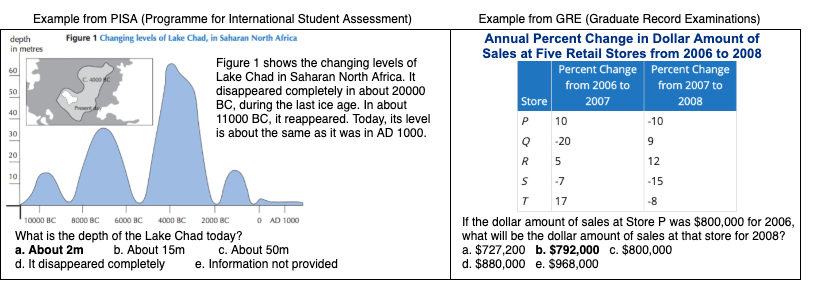}
  \caption{Examples of joint-reasoning questions in standardized tests\footnotemark (boldface represents correct answer)}
  \label{fig:motivation}
\end{figure*}

Question answering (QA) is a crucial way to evaluate the system's ability to understand text and images. In recent years, a large body of natural language QA (NLQA) datasets and visual QA (VQA) datasets have been compiled to evaluate the ability of a system to understand text and images. For most VQA datasets, the text is used merely as a question-answering mechanism rather than an actual modality that provides contextual information. On the other hand, deriving inference from combined visual and textual information is an important skill for humans to perform day-to-day tasks. For example, product assembly using instruction manuals, navigating roads while following street signs, interpreting visual representations (e.g., charts) in various documents such as newspapers and reports, understanding concepts using textbook-style learning, etc. The importance of joint reasoning has also been emphasized in the design of standardized / psychometric tests like PISA \cite{oecd2019pisa} and GRE\footnote{ \url{https://www.oecd.org/pisa/}, \url{https://www.ets.org/gre/}}, as evident from Figure \ref{fig:motivation}. PISA assessments conducted post 2018 take into account ``the evolving nature of reading in digital societies- which requires an ability to compare, contrast and integrate information from multiple sources". The GRE has `data interpretation' questions that assess a student's ability to ``analyze given data as a combination of text and charts." 

 Both the aforementioned evidence motivate the need to develop Visuo-Linguistic QA (VLQA) system, posing a further challenge to state-of-the-art vision and language research. There are no benchmarking datasets that focus on reasoning over both images and text to our best knowledge. 
 \footnotetext{Often, additional text and question are combined in standardized tests, but we segregate them into Passage and Question for the ease of processing and structured dataset design.}
 We formalize the task of deriving joint inference, where a system must utilize both visual and textual information to correctly answer the question, as demonstrated in Figure \ref{fig:coverexample}. To create a benchmark for this task, we develop and present a new dataset: VLQA (Visuo-Linguistic Question Answering)\footnote{Creation of VLQA is purely research-oriented; By referring standardized tests as an inspiration, comparison with professional organizations like ETS or OECD is not intended.} as our main contribution. VLQA dataset consists of text together with a diverse range of visual elements. Since manuals, documents and books containing texts and visuals are ubiquitous, the VLQA dataset is very much grounded in the real world. The dataset is curated from multiple resources (books, encyclopedias, web crawls, existing datasets, etc.) through combined automated and manual efforts. The dataset consists of 9267 image-passage-QA tuples with detailed annotation, which are meticulously crafted to assure its quality. 

We then evaluate the best existing vision-language architectures with respect to our VLQA dataset. This includes LXMERT \cite{tan2019lxmert}, VL-BERT \cite{lu2019vilbert}, ViLBERT \cite{su2019vlbert} and VisualBERT \cite{li2019visualbert}. Our results demonstrate that despite a significant improvement over vision and language tasks separately, the best existing techniques cannot reason well on the joint tasks. We then propose a modular method HOLE (HOpping and Logical Entailment), which demonstrates slightly better baseline performance and offers more transparency for the interpretation of intermediate outputs. The results indicate that VLQA task is relatively harder compared to existing vision-language tasks due to diversity of figures and additional textual component, demanding the need of better approaches to tackle multi-modal question answering. The VLQA challenge thus has the potential to open new research avenues spanning language and vision. 
 
\section{Related Work}
We identify Image-Text Multi-modality, Multi-hop Reasoning and variants of Visual Question Answering (VQA) closest to VLQA and compare with relevant datasets in these areas (refer Appendix \ref{section:survey} for comprehensive comparison with more datasets).  

\subsection{Image-Text Multi-modality}
Multimodal learning aims to build models that can process and relate information from two or more modalities. Image-Text multi-modality has received growing interest from the Artificial Intelligence (AI) community recently. 
Diagram QA component of TQA \cite{kembhavi2017you} and a portion of AI2D \cite{kembhavi2016diagram} with additional text are most relevant to ours. They share similarities with VLQA in terms of the presence of additional text, diagram style images and QA style evaluation, but there are important distinctions. 

 First, TQA uses long lessons ($\sim$50 sentences and 4-5 images) to describe concepts in textbook-style learning, whereas text passages for subsets of AI2D and VLQA are short (1-5 sentences). The goal of TQA aligns with the careful selection of necessary facts from the long-tailed contexts, which is perhaps less important in VLQA as the context is much smaller. At the same time, AI2D aims at AI-based diagram understanding. Contrary to that, we focus on enhancing the capability of AI models for joint reasoning. Secondly, AI2D and TQA are curated from the school science curriculum whereas, we have a broader horizon of possible reasoning. Lastly, TQA and AI2D do not impose that one must use both modalities while answering, unlike VLQA. For TQA, one can answer 40\% of text QA using a single sentence and 50\% of diagram QA using the only image. In that case, a significant portion of the dataset becomes analogous to machine comprehension or ordinary VQA, losing out on the actual purpose of multi-modality.  
\subsection{Multi-Hop Reasoning} 
In the natural language processing (NLP) domain, multi-hop reasoning is proposed to encourage the development of models that can reason about two or more textual contexts. QAngaroo \cite{welbl2018constructing} and ComplexWebQuestions \cite{talmor2018web} include multi-hop questions that can be answered by linking entities from a knowledge base (KB). HotpotQA \cite{yang2018hotpotqa} is a multi-hop benchmark over pairs of text paragraphs from wikipedia, not being constrained by retrieval from fixed KB schemas. QASC \cite{khot2019qasc} dataset made this task further challenging, which first requires to retrieve necessary facts from a large corpus (knowledge ranking) and compose them to answer a multi-hop question. 

Solving VLQA examples requires linking information from image and text. Therefore, VLQA can be considered a novel kind of multi-hop task involving images and text, which we believe will drive future vision-language research.

\subsection{Visual Question Answering (VQA)} 
 Followed by the success of the VQA dataset \cite{antol2015vqa}, several variants of visual QA have been proposed. The following are most relevant;
\paragraph{Reasoning-based VQA} Reasoning-based VQA datasets aim at measuring a system's capability to reason about a set of objects, their attributes and relationships. 
HowManyQA \cite{trott2017interpretable}
and TallyQA \cite{acharya2019tallyqa} have object counting questions over images. SNLI-VE \cite{xie2019visual}, VCOPA \cite{yeo2018visual} focus on causal reasoning whereas CLEVR \cite{johnson2017clevr}, NLVR \cite{suhr2017corpus} target spatial reasoning. 
FigureQA \cite{kahou2017figureqa}, DVQA \cite{kafle2018dvqa} are testbeds for QA over charts/plots.  The objective of VLQA is to equip AI models with diverse reasoning capabilities over the image-text context. A model solving VCR \cite{zellers2019recognition} dataset first answers a question in VQA style, then needs to provide a rationale explaining why the answer is true. Therefore, items in VCR could be turned to particular VLQA data items. However, images in VCR are much more specific than ours e.g., they do not have charts, diagrams, or multiple images. Also, the rationale selection is limited to `Why' questions, not so in VLQA.  We identify 10 broad reasoning categories needed to solve VLQA, which is described in Section  \ref{sec:analysis}. 

\paragraph{Knowledge-based VQA} There are several vision-language tasks that require additional knowledge beyond the provided image and text. F-VQA \cite{wang2018fvqa}, KB-VQA \cite{wang2015explicit} and  KVQA \cite{shah2019kvqa} rely on retrieving commonsense or world-knowledge from a Knowledge Base (KB), whereas OK-VQA \cite{marino2019ok} is related to open-ended knowledge extraction from the web. In VLQA, 61\% of samples require commonsense or domain knowledge, which is not explicitly stated in image-text context. Knowledge extraction for VLQA is kept open-ended as of now. 

\section{VLQA Dataset}

We formally define the VLQA task, explain our approach to curate this dataset and necessary measures for quality assurance below;

\subsection{Task Overview}

A datapoint in VLQA is a 4-tuple $<$I, P, Q, A$>$;

\paragraph{Image(I)} It is provided imagery, which ranges from daily life scenes, a variety of data representations to complex diagrams. A portion of VLQA examples also requires reasoning over multiple images. For the simplicity of processing and retrieval, we compose all images into a single file. Each image is bounded by a red box and provided an explicit detection tag ([0],[1],..) for identification purposes, inspired by VCR \cite{zellers2019recognition} annotations. This also provides a convenient way to reference images in passage, question, or answers.
 
 \begin{figure*}
  \center
 \tab Example 1 \tab Example 2 \tab Example 3 \\
  \includegraphics[width=\textwidth]{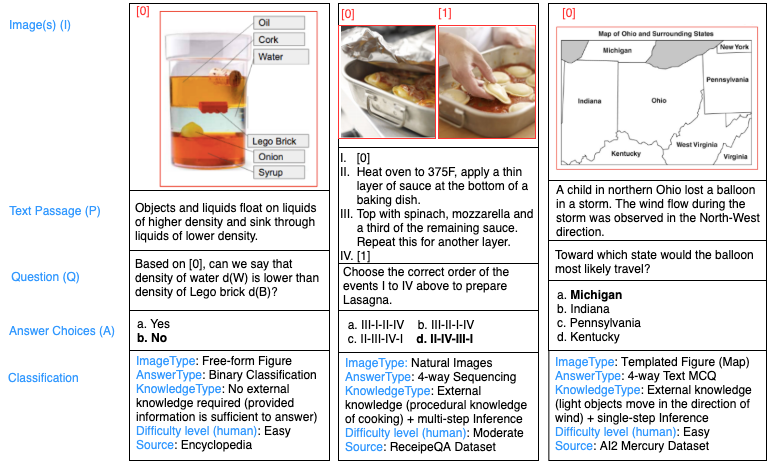}
  \caption{\textbf{Examples from VLQA Train Set}. Each example contains image, corresponding text passage and Multiple Choice Question (MCQ) with correct answer choice highlighted by the boldface. Further, each sample is classified based on image type, answer type, knowledge/reasoning type and human annotated difficulty level.\\ (For more examples, refer to \ref{sec:moreex} or visit dataset webpage
  }
  
  \label{fig:example}
\end{figure*} 

\paragraph{Passage(P)} It is a textual modality that provides additional contextual information related to the image. The passages in VLQA dataset is composed of 1-5 sentences, which consists of facts, imaginary scenarios or their combination.

\paragraph{Question(Q)} It is a question in natural language that tests the reasoning capability of a model over a given image-passage context. In addition to standard `Wh' patterns and fact-checking style (True/False), some questions in VLQA are of `do-as-directed' form, similar to standardized tests. 

\paragraph{Answer Choices(A)} VLQA is formed as a classification task over 2-way or 4-way plausible choices, with exactly one of the candidate answers being correct. Answer choices may contain boolean, alpha-numeric phrases, image tags or their combination. 

\paragraph{Task} Given the VLQA dataset as a collection of 4-tuple $<$I, P, Q, A$>$ as shown in Figure \ref{fig:example}, the task is to build an AI model that can answer a given question using image-text multi-modal context. The correctness of the prediction is measured against the ground-truth answer. Additionally, we provide rich annotations and classification on several aspects such as image types, question types, required reasoning capability and need for external knowledge. However, this metadata is optional and useful for researchers interested in tackling specific subsets of VLQA.

\subsection{Constructing VLQA}

\subsubsection{Data Collection} 

The main goal of our work is to collect a QA dataset that requires to derive joint inference from image-text modality. We classify our data sources as Primary and Secondary;

We obtain raw textual/visual information through primary sources, which can be later used as a modality in VLQA. For example, text crawls from wikipedia containing facts or images crawled by keyword-search can be used as passage and image respectively. Similarly, we collect tabular data from CIA `world factbook' \cite{cia2019factbook}, WikiTables \cite{pasupat2015compositional} and convert them into templated figures like bar charts, pie charts, scatter plots, etc.
\begin{figure*}
  \center
  \includegraphics[width=\textwidth]{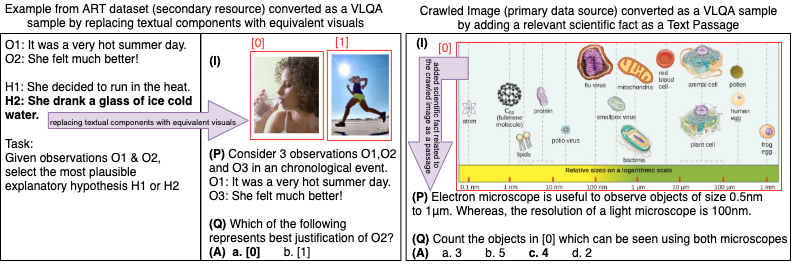}
  \caption{VLQA data creation process: collect data using primary and secondary sources, then perform post-processing (if any), then finally create question-answers that require joint reasoning.
  }
  \label{fig:example2}
\end{figure*}
We consider existing structured or semi-structured materials as a secondary data source, which can be quickly manipulated to use for our purpose; educational materials, standardized tests, and existing vision-language datasets are important.  We used scrapers to collect textbook exercises, encyclopedias,  practice worksheets and question banks. Further, we obtained a subset of interesting samples from existing datasets such as RecipeQA \cite{yagcioglu2018recipeqa}, WikiHow \cite{koupaee2018wikihow}, PhysicalIQA \cite{bisk2019piqa}, ART \cite{ch2019abductive} and TQA \cite{kembhavi2017you}. 

We then refactor textual/ visual information collected from the above sources and mold it as per our task requirements. Figure \ref{fig:example2} illustrates this process. Refactoring includes manual or semi-automated post-processing such as replacing given textual/visual attributes with equivalent visual/textual counterparts, adding/removing partial information to/from text or visuals, and creating factual or hypothetical situations around images. 
Then we standardize all information collected the using above methods as Multiple Choice Questions (MCQ) and get the initial version of the dataset. 

Since we impose the condition that a question must be answered through joint reasoning over both the modalities, our annotation process becomes non-trivial and requires careful manual annotation. We opted for a limited number of in-house expert annotators for quality purposes rather than a noisier hard-to-control crowdsourcing alternative.  

\subsubsection{Ensuring dataset integrity}
A combined understanding about visual and textual inputs is a key aspect of the VLQA task. As we model it as a classification task, some models might exploit various biases in the dataset to get good performance without proper reasoning. To discourage such models, we employ 3-level verification over the full dataset to ensure the quality.

Firstly, for all collected image-passage pairs, human annotators quickly verify if a portion of image and passage represent identical information. All such image-passage pairs are discarded from the dataset. Secondly, we create 3 baselines- question-only, passage-only and image-only which ignore at least one modality (among image and passage) and try to predict answers. We repeat this experiment 3 times by shuffling answer choices with a fixed seed. We remove samples that are answered correctly by any unimodal baseline in all trials. 

Finally, we perform another round of manual quality checks. We instruct workers first to answer a question based only on image(s) and then try to answer a question based only on the text passage. If a question can be answered using a single modality, we suggest annotators to mark the checkbox. Finally, we look over all bad samples and either provide a fix or remove, on a case-by-case basis. (refer Appendix \ref{sec:pipeline} for detailed explanation on dataset creation process)

\subsection{VLQA Dataset Analysis} 
\label{sec:analysis}
In this section we analyze VLQA on following aspects; Table \ref{tab:tab1} provides a summary of relevant statistics. 

\paragraph{Multi-modal Contexts} The final version of the VLQA dataset has 9267 unique image-passage-QA items. For each item, the multi-modal context is created by pairing images (roughly 10k collected) with the relevant text passages (roughly 9k retrieved or manually written).

\paragraph{Text-length Analysis}
We provide analysis about lengths of various textual components in our dataset i.e., passages, questions and answers. Length of each textual component is calculated by counting the tokens separated by whitespaces and then averaged out across the dataset. The average passage length of 34.1 tokens indicates that in VLQA textual contexts are relatively smaller than Reading Comprehension tasks and in most cases, it contains precise context necessary for the joint reasoning. The average question length of 10.0 tokens is larger compared to most other VQA datasets provided in \cite{hudson2019gqa}. Shorter answer lengths (1.7 tokens) suggest that most of the dataset questions have short answers, which provides inherent flexibility if someone wants to leverage generative models to solve this task. The dataset has a vocabulary size of 13259, contributed by all three textual components together.

\paragraph{Image types} 
 We categorize images in VLQA into 3 major kinds: Natural Images, Template-based Figures and Free-form Figures. Natural images incorporate day-to-day scenes around us, containing abundant objects and actions. Template-based figures are visuals that follow a common structure for information representation. We further categorize template-based figures into 20 sub-types like bar, pie, maps, tables, cycles, processes, etc. The images which neither fit in any templates nor are natural have been put into a free-form category (e.g., science experiments, hypothetical scenarios, etc.). In VLQA, it is also possible that the visual context has multiple related images to reason about.  
 
\paragraph{Answer types} 
4-way or 2-way image MCQ contains 4 and 2 images as plausible answer choices respectively, where the model needs to correctly pick the image best described by the passage and question. 4-way or 2-way text MCQ contains 4 and 2 alphanumeric text as plausible answer choices respectively, where the model needs to reason about given image-text scenario and pick the most likely answer to the question. 4-way Sequencing task assesses a model's capability to order 4 spatial or temporal events represented as a combination of images and text. Binary Classification (Yes/No or True/False) can be considered a fact-checking task where we want to determine the truth value of a question provided image-passage context.

\paragraph{Knowledge and Reasoning types} 
 61\% of VLQA items are observed to incorporate some commonsense or domain knowledge beyond the provided context. This missing knowledge has to be retrieved through the web. The remaining 39\% samples can be answered through a simple join of information from visuo-linguistic context. We observe the following 10 most-frequent reasoning types needed to solve VLQA questions; conditional retrieval, math operations, deduction, temporal, spatial, causal, abductive, logical, and verbal reasoning. We further categorize VLQA samples based on whether it requires a single-step or multi-step inference to answer the question. By multi-step inference, we mean that answering a question involves more than one reasoning types. 

\renewcommand{\tabcolsep}{3pt}
\begin{table}[H]
\centering
\begin{tabular}{llrr}  
\toprule
          &         \textbf{Measure}                       &  \textbf{Stats.} 
            \\ 
            \midrule
            
\multicolumn{2}{l}{\textbf{Multimodal Context}} &         \\           
            & Total \#Images             &    10209    \\         
            & \#Unique Text Passages                 &  9156  \\          
            & \#Questions                &   9267   \\          
\midrule
\multicolumn{2}{l}{\textbf{Text-length Analysis}} &         \\           
            & Avg. Passage Length               &    34.1     \\         
            & Avg. Question Length                 &  10.0 \\          
            & Avg. Answer Length                &   1.7   \\          
            & Vocabulary Size                &   13259    \\          
            \midrule
            \multicolumn{2}{l}{\textbf{Image types}} &         \\           
            & Natural Images                 &    4445    \\         
            & Templated Figures                 &  3920   \\          
            & Free-form Figures                &   1854  \\          
             \midrule
             \multicolumn{2}{l}{\textbf{Answer types}} &         \\           
            & 4-way image MCQ                &   1172            
            \\
            & 4-way text MCQ                 &     4647    \\         
             & 4-way Sequencing             &  1088    \\     
            
            & 2-way image MCQ                &    1088   \\       
             & Binary Classification (T/F or Yes/No)              &  1272    \\    
            \midrule
            \multicolumn{2}{l}{\textbf{Knowledge/Reasoning types}} &         \\           
            & No Ext. Knowledge required
                &   3145         
            \\
             & Ext. Knowledge+Single-step Inference 
                 &  2783    \\    
            & Ext. Knowledge+Multi-step Inference
      &     2939       \\
      \midrule
            \multicolumn{2}{l}{\textbf{Difficulty Level (human annotated)}} & 
            \\           
            & Easy
                &    4188    
            \\
            &  Moderate             &    2943    \\         
             & Hard 
                 &  2136    \\  
    \midrule
            \multicolumn{2}{l}{\textbf{Dataset Split}} & 
            \\           
            & Train (80\%)
                &    7413    
            \\
            &  Test (10\%)             &    927    \\         
             & Validation (10\%)
                 &   927   \\    
     \bottomrule
\end{tabular}
\caption {\label{tab:tab1} VLQA Statistics and Diversity (MCQ is multiple choice questions, Ext. is External). 
}
\end{table} 

 \paragraph{Difficulty Level}
Determining difficulty levels is a subjective notion therefore, we asked an odd number of annotators to rate VLQA items as `easy', `moderate', or `hard' based on their personal opinion. Then we take a majority vote of all annotators to assign difficulty level to each question.

\paragraph{Dataset Splits}
VLQA contains 9267 items in $<$I,P,Q,A$>$ format, with detailed classification based on figure types, answer types, reasoning skills, requirement of external knowledge and difficulty levels as explained above. The data is split in train-test-val (80-10-10\%), ensuring the uniform distribution based on the above taxonomies. To preserve the integrity of the test results, we do not release the test set publicly. Note that the use of the metadata for model design is completely optional. 

\section{Benchmarking}

\paragraph{Human Performance} 
 We performed human evaluation on 927 test samples with a balanced variety of questions by image types, answer types, knowledge/reasoning types and hardness. First, we ask 3 in-house experts to take tests in isolation. We also ask them to rate questions based on the difficulty levels (easy/medium/hard) and an option to mark a dataset sample `ambiguous'. Then we match their predictions against ground-truth answers, which turned out to be 84\%. 
 
\begin{figure*}
  \center
  \includegraphics[width=\textwidth]{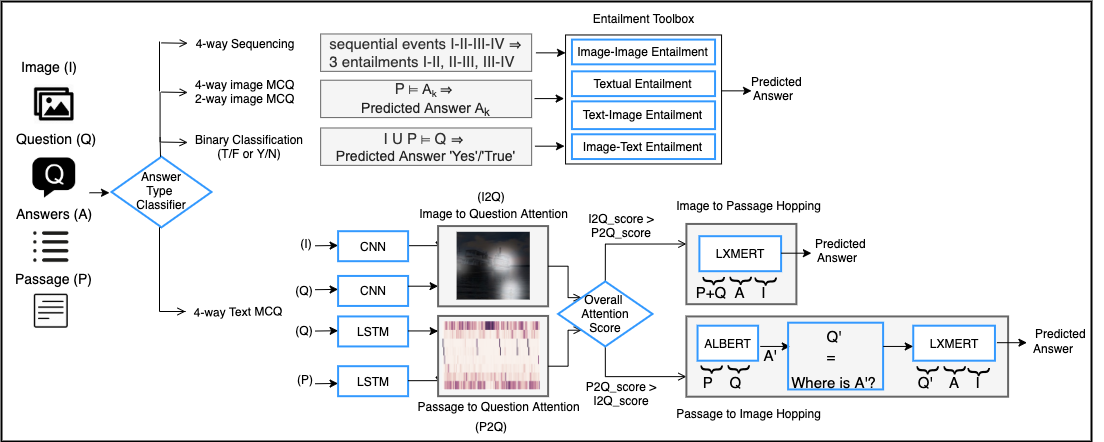}
  \caption{Proposed HOLE method to solve VLQA: Based on the answer type classification, a dataset item is solved as a sequence of Logical Entailment operations or performs Hopping between modalities to find the correct answer.
  }
  \label{fig:archi}
\end{figure*}

\paragraph{Random Baseline} VLQA dataset contains 4-way and 2-way multiple choice questions (MCQs) where each answer choice is likely to be picked with 25\% and 50\% chance. Based on the answer-type distribution provided in Table \ref{tab:tab1}, the performance of the random baseline is 31.36\%. 

\paragraph{Question-only, Passage-only and Image-only Baselines}
We use three unimodal baselines only for automated quality assurance of VLQA data (and do no not train) to prevent models from exploiting bias in data. Question-only, Passage-only and Image-only models are implemented using RoBERTa \cite{liu2019roberta} finetuned on ARC \cite{clark2018think}, ALBERT \cite{lan2019albert} finetuned on RACE and LXMERT \cite{tan2019lxmert} finetuned on VQA \cite{antol2015vqa} respectively. We report the poor performance of these baselines over resulting VLQA data to indicate the need for joint reasoning over multi-modal context.

\paragraph{Best Existing Architectures} Recently, several attempts have been made to derive transformer-based pre-trainable generic representations for visuo-linguistic tasks. We pick top-performing single-model architectures VL-BERT \cite{su2019vlbert}, VisualBERT \cite{li2019visualbert}, ViLBERT \cite{lu2019vilbert} and LXMERT \cite{tan2019lxmert} that support Visual Question Answering (VQA) downstream task. For the VQA task, the input is an image and a question. To finetune VQA style models with VLQA data, we compose all images into one (in case of multiple images) as a single visual input, and concatenate Passage and Question as a single language input. Hyperparameters and Performance of all 4 architectures is reported in \ref{tab:tab3} and \ref{tab:tab2} respectively.

\begin{table}[h]
\begin{tabular}{@{}l@{}}
\toprule
\multicolumn{1}{c}{\textbf{Model and Hyperparameters}}                                                                                \\ \midrule
\begin{tabular}[c]{@{}l@{}}\textbf{VisualBERT}\\  $\frac{}{}\frac{}{}$  Ft\_VQA: EP=20, BS=256, LR=1e-4, WD=1e-4\\  $\frac{}{}\frac{}{}$   Ft\_VLQA: BS=16, LR=2e-5, EP=15\end{tabular} \\ \midrule
\begin{tabular}[c]{@{}l@{}}\textbf{VL-BERT }   \\ $\frac{}{}\frac{}{}$    Ft\_VQA: BS=32, LR=2e-5, EP=10\\ $\frac{}{}\frac{}{}$    Ft\_VLQA: BS=16, LR=1e-5, EP=10\end{tabular}          \\ \midrule
\begin{tabular}[c]{@{}l@{}}\textbf{ViLBERT}\\  $\frac{}{}\frac{}{}$   Ft\_VQA: BS=32, LR=1e-5, EP=20, WR=0.1\\  $\frac{}{}\frac{}{}$   Ft\_VLQA: BS=32, LR=1e-5, EP=10\end{tabular}      \\ \midrule
\begin{tabular}[c]{@{}l@{}}\textbf{LXMERT}\\  $\frac{}{}\frac{}{}$   Ft\_VQA: BS=32, LR=5e-5, EP=4\\  $\frac{}{}\frac{}{}$   Ft\_VLQA: BS=16, LR=5e-5, EP=8\end{tabular}                 \\ \bottomrule
\end{tabular}
\caption{Manual finetuning of best existing architures with VQA followed by VLQA (BS-Batch Size, EP- Epochs, LR-Learning Rate, WD-Weight Decay, WR-Warmup Ratio, Ft.-Manual Finetuning)} 
\label{tab:tab3}
\end{table}

\section{Fusion of HOpping and Logical Entailment (HOLE) to solve VLQA}

We propose `HOLE'- a fusion of modality HOpping (Image-to-passage hop and Passage-to-Image hop) and Logical Entailment as a modular baseline for VLQA, shown in Figure \ref{fig:archi}. We leverage `answer types' metadata from the annotations and learn a simple 5-class classifier (`4-way Image', `2-way Image', `4-way Sequencing', `Binary Classification' or `4-way Text') in order to decide between modality hopping and logical entailment. Note that our model is not end-to-end. 
\subsection{Modality Hopping based Solver}

\textbf{4-way text MCQ} are solved using modality hopping approach (lower half pipeline in Figure \ref{fig:archi}). We first compute Image-to-Question Attention (I2Q) and Passage-to-Question Attention (P2Q) scores to determine which modality is important as a starting point for solving a question. I2Q is computed using Stacked Attention Network (SAN) \cite{yang2016stacked}, which takes Convolution Neural Network (CNN) encoding of I and Q. Whereas, P2Q is computed using a variant of Bi-Directional Attention Flow (BIDAF) \cite{seo2016bidirectional} trained using Embeddings from Language Models (ELMo) \cite{Peters:2018} over Long-Short Term Memory (LSTM) encoding of Q and P. 

A higher I2Q score suggests that Q has more overlap with I than P. Therefore, image modality should be used first and then incorporate passage to compute the answer. This is termed as an `Image-to-Passage Hop'. This is identical to a Visual Question Answering (VQA) scenario that takes an image and a question as input. Since we have P as an additional text component, we combine passage (P+Q). This is implemented through pre-trained architecture LXMERT \cite{tan2019lxmert} which is state-of-the-art on VQA that picks the most likely answer choice as a correct answer. 

Similarly, a higher P2Q score suggests that Q has more overlap with P than I. Therefore, passage modality should be used first and then incorporate image to compute the answer. This is termed as a `Passage-to-Image Hop'. This can be achieved by a machine comprehension model followed by a VQA model. We use ALBERT \cite{lan2019albert} as a machine comprehension model which takes in P and Q to generate an open-ended response in the style of SQuAD \cite{rajpurkar2016squad}, which we refer to as A'. Now we want to determine where is A' located in the image I. Therefore, we formulate a new question Q' as ``Where is A'?'', where A' is substituted by the answer from ALBERT. We then use LXMERT \cite{tan2019lxmert} that takes image I, new question Q' and original answer choices A to pick the most likely one.  
\subsection{Logical Entailment based Reasoner}\footnote{$\vdash$ is the symbolic representation of  entailment}
For all other answer types, we leverage Logical Entailment (upper half pipeline in \ref{fig:archi}) of image and text to answer questions. We create an `Entailment Toolbox' which consists of image-image, image-text \cite{xie2019visual}, text-image and textual entailment \cite{scitail} sub-modules and use them as required. 
For image-image and image-text entailment, we augment Visual COPA \cite{yeo2018visual} dataset and train custom network for both.
(refer Supplementary Material \ref{sec:supplemental} for more details)

\paragraph{4-way or 2-way image MCQ} contains images as an answer choice, which is similar to an Image Selection task \cite{Hu_2019_ICCV}. The goal here is to identify an image that best matches the description of P or mathematically, determine $P \vdash A_k$ (i.e., text-image entailment) with maximum score. $A_k$ represents answer choices where k=4 and k=2 for 4-way and 2-way image problems respectively.

\paragraph{Binary Classification} can be considered as a fact-checking task where we want to determine the truth value of a question provided image-passage, or mathematically, $P \cup I \vdash Q $. We use textual entailment to determine $P\vdash Q$ and image-text entailment to determine $I\vdash Q$. If both entailment modules' confidence score is above 0.65 then it is determined as True, otherwise False.  

\paragraph{4-way Sequencing} task assesses a model's capability to order 4 spatial or temporal events. If we consider I-II-III-IV as a sequence of events, it is equivalent to 3 entailment tasks: I-II, II-III, and III-IV, where each I to IV can be an image or a text. Among the answer choices, the sequence with maximum overall confidence is selected as an answer. 

\section{Results \& Discussion}

Multi-modality brings both pros and cons while developing new Artificial Intelligence (AI) benchmarks. The presence of multiple modalities provide natural flexibility for varied inference tasks, simultaneously making the reasoning process more complex as information is now spanned across them and requires cross-inferencing. In this work, we focused on joint reasoning over image-text multi-modal context and developed a Visuo-Linguistic Question Answering (VLQA) Dataset. Our proposed VLQA dataset has important distinctions from existing VQA datasets. Firstly, it incorporates a text passage that contains additional contextual information. Secondly, it offers various figure types including natural images, templated images and free-form images (unstructured), which is not so common for other VQA datasets. Thirdly, it tests diverse reasoning capabilities, including cross-inferencing between visual and textual modalities. 

We then use several baselines and benchmark their performance over the resulting VLQA dataset. As VLQA has multiple choice questions with exactly one correct answer, we use standard accuracy as an evaluation metric. From the results in \ref{tab:tab2}, we can observe that pre-trained vision-language models fail to solve a significant portion of the VLQA items. Our proposed modular method HOLE slightly outperforms them and is more interpretable for analysis. We also report the performance of Question-only, Image-only and Passage-only baselines which we used for quality check. The poor performance of these baselines indicate that the VLQA dataset requires models to jointly understand both image and text modalities and is relatively harder than other  vision-language tasks. 

For human evaluation of the VLQA test-set, the reported accuracy is 84.0\%. For 148 wrongly predicted answers, we group them according to 4 reasons for failures, which are listed in \ref{tab:tab4}. The results demonstrate a room for significant improvement in existing vision-language models that are far behind the human performance. This stimulates the need for more complex reasoning capabilities of AI models. We suspect that VLQA questions that purely rely on facts might be exploited by the latest language models, despite strong measures taken through manual and automated quality control during the creation of the dataset. We would like to explore this further in the future.

\renewcommand{\tabcolsep}{4pt}
\begin{table}
\centering
\begin{tabular}{llrr}  
\toprule
\multicolumn{2}{c}{\textbf{Method}}                    &  \multicolumn{1}{c}{\textbf{Test(\%)}} & \multicolumn{1}{c}{\textbf{Val(\%)}} \\
\midrule
\multicolumn{2}{l}{Human} &        84.00            &    --     \\
\midrule
\multicolumn{2}{l}{Random}                &       31.36      &     31.36            \\
\midrule
\multicolumn{2}{l}{Question-only: \small{RoBERTa$_{ARC}$}}  &    28.56       &   29.42     \\
\multicolumn{2}{l}{Passage-only:  \small{ALBERT$_{RACE}$}}  &       30.16     &   30.25   \\
\multicolumn{2}{l}{Image-only: \small{LXMERT$_{VQA}$}}   &   29.48         &   30.56     \\
\midrule
\multicolumn{2}{l}{Vision-Language} &                                   \\
& VL-BERT                             &       35.92      &   34.60            \\
         & VisualBERT       &         33.17   &    34.17          \\
         & ViLBERT             &     34.70    &     35.25         \\
          & LXMERT          &      \textbf{36.41}   &     37.82             \\  
          
\midrule
\multicolumn{2}{l}{HOLE (Proposed Model)}    &   \textbf{39.63}    &   40.08        \\
\bottomrule
\end{tabular}
\caption{\label{tab:tab2} Performance benchmarks over test-set of VLQA task and corresponding validation results}
\end{table}

\begin{table}[]
\begin{tabular}{@{}ll@{}}
\toprule
\begin{tabular}[c]{@{}l@{}}\textbf{Underlying reason for incorrect} \\ \textbf{answer provided by test-taker} \end{tabular}  & \begin{tabular}[c]{@{}l@{}}\textbf{\#incorrect/148} \\ \textbf{(\%incorrect)} \end{tabular}  \\ \midrule
\begin{tabular}[c]{@{}l@{}}Lacked necessary knowledge  \end{tabular}  & \begin{tabular}[c]{@{}l@{}}27 (18.2\%)\end{tabular} \\
\begin{tabular}[c]{@{}l@{}}Misunderstood the provided info \end{tabular}  & \begin{tabular}[c]{@{}l@{}}47 (31.7\%)\end{tabular} \\
\begin{tabular}[c]{@{}l@{}}Mistake in deduction/calculation \end{tabular}  & \begin{tabular}[c]{@{}l@{}}63 (42.5\%)\end{tabular} \\
\begin{tabular}[c]{@{}l@{}}Felt that data item is ambiguous \end{tabular}  & \begin{tabular}[c]{@{}l@{}}11 (7.4\%)\end{tabular} \\ 
\bottomrule
\end{tabular}
\caption{\label{tab:tab4} Classification of incorrectly predicted answers in Human-evaluation of VLQA test-data}
\end{table}

\section{Conclusion}

In this work, we introduced the Visuo-Linguistic Question Answering (VLQA) challenge that we believe has the potential to open new research avenues in areas of joint vision \& language. Our experiments show that a system equipped with state-of-the-art vision-language pre-training does not perform well on the task that requires joint image-text inference. There is a room for significant improvement in capability of these models to tackle multi-modal contexts. Our future work would include further expansion of this dataset and building generic AI models that can learn novel visual concepts from a small set of examples.

\section*{Acknowledgments}

We are thankful to the anonymous reviewers for the feedback. This work is partially supported by the National Science Foundation grant IIS-1816039.

\bibliographystyle{acl_natbib}
\bibliography{emnlp2020}

\appendix

\clearpage
\section{Appendices}
\label{sec:appendix}

\renewcommand{\tabcolsep}{2pt}
\begin{table*}
\begin{tabular}{@{}llllllllll@{}}
\toprule
\textbf{Dataset} & \multicolumn{4}{l}{\textbf{Modality}} & \multicolumn{2}{l}{\textbf{Modality Classification}} & \textbf{Task Type} & \textbf{Task (Domain)} \\ \midrule
 & \textbf{I} & \textbf{T} & \textbf{T+} & \textbf{K} & \textbf{Visual} & \textbf{Textual} &  &  \\ \midrule
\href{https://arxiv.org/pdf/1612.06890.pdf}{Clevr} & \cmark & \cmark & \xmark & \xmark & Synthetic & Ques & OE & VQA (Spatial Reasoning) \\
\href{https://arxiv.org/abs/1405.0312}{COCO} & \cmark & \cmark & \xmark & \xmark & Natural & Caption & Caption & Text generation  \\
\href{https://arxiv.org/abs/1901.06595}{COCO-BISON} & \cmark  & \cmark & \xmark & \xmark & Natural & Sent & MC & Image Selection \\
\href{https://arxiv.org/pdf/1505.02074.pdf}{COCO-QA} & \cmark  & \cmark & \xmark & \xmark & Natural & Ques & OE & VQA \\
\href{https://arxiv.org/abs/1803.06092}{COG} & \cmark &  \cmark & \xmark & \xmark & Synthetic & Ques / Sent & MC & VQA, Instruction Following \\
\href{https://www.aclweb.org/anthology/P18-1238.pdf}{Concept.Caption} & \cmark  & \cmark & \xmark & \xmark & Natural & Caption & Caption & Text generation \\
\href{http://openaccess.thecvf.com/content_cvpr_2017/papers/Chattopadhyay_Counting_Everyday_Objects_CVPR_2017_paper.pdf}{CountQA} & \cmark & \cmark & \xmark & \xmark & Natural & Ques & Numeral & VQA (Counting) \\
\href{https://papers.nips.cc/paper/5411-a-multi-world-approach-to-question-answering-about-real-world-scenes-based-on-uncertain-input.pdf}{DAQUAR} & \cmark  & \cmark & \xmark & \xmark & Natural & Ques & OE & VQA \\
\href{https://arxiv.org/pdf/1801.08163.pdf}{DVQA} & \cmark  & \cmark & \xmark & \xmark & Synthetic & Ques & OE & VQA (BarCharts) \\
\href{https://arxiv.org/abs/1710.07300}{FigureQA} & \cmark & \cmark & \xmark & \xmark & Synthetic & Ques & OE & VQA (Charts) \\
\href{http://papers.nips.cc/paper/5641-are-you-talking-to-a-machine-dataset-and-methods-for-multilingual-image-question.pdf}{FMIQA} & \cmark & \cmark & \xmark & \xmark & Natural & Ques & OE & VQA \\
\href{https://arxiv.org/pdf/1902.09506.pdf}{GQA} & \cmark  & \cmark & \xmark & \xmark & Natural & Ques & OE & VQA \\
\href{https://arxiv.org/abs/1712.08697}{HowManyQA} & \cmark  & \cmark & \xmark & \xmark & Natural & Ques & Numeral & VQA (Counting) \\

\href{https://arxiv.org/abs/1907.12861}{LEAFQA} & \cmark  & \cmark & \xmark & \xmark & Synthetic & Ques & OE & VQA (Charts) \\
\href{https://ieeexplore.ieee.org/stamp/stamp.jsp?tp=&arnumber=8603827}{Memex-QA} & \cmark  & \cmark & \xmark & \xmark & Natural & Ques & MC & VQA \\
\href{https://www.microsoft.com/en-us/research/wp-content/uploads/2016/06/cvpr16.msr-vtt.tmei_-1.pdf}{MSRVTT-QA} & \cmark  & \cmark & \xmark & \xmark & Natural & Ques & OE & VQA \\
\href{https://yoavartzi.com/pub/slya-acl.2017.pdf}{NLVRv1/v2} & \cmark  & \cmark & \xmark & \xmark & Synthetic/Natural & Sent & T/F & Text classification \\
\href{https://arxiv.org/pdf/1811.00982.pdf}{OpenImagesV6} & \cmark  & \cmark & \xmark & \xmark & Natural & Caption & Caption & Text generation\\
\href{https://arxiv.org/pdf/1805.09701.pdf}{RVQA} & \cmark  & \cmark & \xmark & \xmark & Natural & Ques & OE & VQA \\
\href{https://pdfs.semanticscholar.org/0ac8/f1a3c679b90d22c1f840cdc8d61ffef750ac.pdf}{Shapes} & \cmark & \cmark & \xmark  & \xmark & Synthetic & Ques & OE & VQA \\
\href{https://arxiv.org/abs/1704.04517}{ShapeWorld} & \cmark & \cmark & \xmark & \xmark & Synthetic & Sent & Scoring & Text classification \\
\href{https://arxiv.org/abs/1901.06706}{SNLI-VE} & \cmark  & \cmark & \xmark & \xmark & Natural & Sent & 3 classes & Visual Entailment \\
\href{https://arxiv.org/pdf/1810.12440.pdf}{TallyQA} & \cmark  & \cmark & \xmark & \xmark & Natural & Ques & Numeric & VQA (Counting) \\
\href{https://arxiv.org/pdf/1703.09684.pdf}{TDIUC} & \cmark  & \cmark & \xmark & \xmark & Natural & Ques & OE & VQA \\
\href{https://arxiv.org/abs/1904.08920}{TextVQA} & \cmark  & \cmark & \xmark & \xmark & Natural & Ques & OE & VQA (Text in Images) \\

\href{https://arxiv.org/pdf/1811.10830.pdf}{VCR} & \cmark & \cmark & \xmark & \xmark & Natural & Ques & MC & VQA+Rationale \\
\href{https://visualgenome.org/static/paper/Visual_Genome.pdf}{Vis.Genome} & \cmark  & \cmark & \xmark & \xmark & Natural & Ques & OE & VQA (Scene Graphs) \\
\href{https://arxiv.org/pdf/1506.00278.pdf}{Vis.Madlibs} & \cmark  & \cmark & \xmark & \xmark & Natural & Sent & Blanks & VQA \\
\href{https://arxiv.org/pdf/1511.03416.pdf}{Vis.7W} & \cmark  & \cmark & \xmark & \xmark & Natural & Ques & MC & VQA \\
\href{https://arxiv.org/abs/1611.08669}{Vis.Dialogue} & \cmark  & \cmark & \xmark & \xmark & Natural & Ques & OE & VQA (Dialogue) \\
\href{https://www.ischool.utexas.edu/~dannag/publications/CVPR2019_VizWiz-Priv.pdf}{VizWiz-Priv} & \cmark &  \cmark & \xmark & \xmark & Natural & Ques & OE & VQA (Text in Images) \\
\href{https://arxiv.org/pdf/1505.00468.pdf}{VQAv1 Abs./Real} & \cmark  & \cmark & \xmark & \xmark & Synthetic/Natural & Ques & OE & VQA \\
\href{https://arxiv.org/pdf/1612.00837.pdf}{VQAv2/CP} & \cmark &  \cmark & \xmark & \xmark & Natural & Ques & OE,MC & VQA \\
\href{https://www.aclweb.org/anthology/D19-5224.pdf}{WAT2019} & \cmark  & \cmark & \xmark & \xmark & Natural & Caption & Caption & Text generation / Translation \\
\midrule
AI2 Geometry & \cmark & \cmark & \xmark & \cmark & Diagrams & Ques & MC & VQA (Geometry) \\
AI2 Mercury & \cmark  & \cmark & \xmark & \cmark & Diagrams & Ques & MC & VQA (Science) \\
AI2 ScienceQ & \cmark  & \cmark & \xmark & \cmark & Diagrams & Ques & MC & VQA (Science) \\
\href{https://arxiv.org/abs/1603.07396}{AI2D} & \cmark &  \cmark & \xmark & \cmark & Diagrams & Ques & MC & VQA (Science) \\
\href{https://arxiv.org/pdf/1606.05433.pdf}{FVQA} & \cmark  & \cmark & \xmark & \cmark  & Natural & Ques & OE & VQA (Commonsense) \\
\href{https://www.ijcai.org/proceedings/2017/0179.pdf}{KBVQA} & \cmark  & \cmark & \xmark & \cmark  & Natural & Ques & OE & VQA (Commonsense) \\
\href{http://dosa.cds.iisc.ac.in/kvqa/KVQA-AAAI2019.pdf}{KVQA} & \cmark  & \cmark & \xmark & \cmark & Natural & Ques & OE & VQA (World Knowledge) \\
\href{https://arxiv.org/pdf/1906.00067.pdf}{OKVQA} & \cmark  & \cmark & \xmark & \cmark & Natural & Ques & OE & VQA (World Knowledge) \\
\href{https://github.com/sanket0211/WK-VQA/issues/1}{WKVQA} & \cmark  & \cmark & \xmark & \cmark & Natural & Ques & OE & VQA (World Knowledge) \\
\midrule
 \href{http://openaccess.thecvf.com/content_cvpr_2018/papers/Li_Textbook_Question_Answering_CVPR_2018_paper.pdf}{TQA} & \cmark  & \cmark & \cmark & \cmark  & Diagrams & Ques, Lesson & MC & VQA (Science) \\
\midrule
{VLQA (Our)} & \cmark  & \cmark & \cmark & \cmark  & Natural, & Ques, Para & MC & VQA (Joint Reasoning \\
 &  &  & & & Synthetic, & & & over Image-Text) \\
 &  &  &  &  & Diagrams  & & & 
 \\

\bottomrule
\end{tabular}
\caption{\label{tab:tab5} Survey of existing vision-Language Datasets and Comparison with VLQA}
\end{table*}

\subsection{Survey of existing vision-Language Datasets and Comparison with VLQA}
\label{section:survey}
There are several datasets proposed in recent years to benchmark a variety of vision-language tasks. We provide the list of such datasets and  comparison with our dataset in Table \ref{tab:tab5} based on following attributes; 
\begin{enumerate}
    \item \textbf{Dataset} name with corresponding url of dataset website/publication is available. 
    \item \textbf{Modality} states which of the following components each dataset has; I (Images), T (Text as a QA mechanism), T$+$ (Additional Textual Context), K (Additional Knowledge required). VLQA dataset has all four components, standing out from the rest of the datasets except TQA, which has a different objective of textbook-style learning. This makes VLQA task harder than other existing datasets, which we believe will be a driver for development of more advanced AI models.    
    \item \textbf{Visual Modality Classification} describes the nature of visuals incorporated for a dataset which are categorized in 3 major kinds; Natural (everyday objects and scenes), Synthetic (artificially/program generated or templated figures) or Diagrams (imagery representing complex relationships between multiple interrelated objects or phenomena). Our dataset includes all three kinds of visuals aiming at developing generic vision-language reasoning system. However, we provide this classification as a part of our annotations for researchers interested in advancements specific to a particular kind of visual. 
    \item \textbf{Textual Modality Classification} describes the nature of language component incorporated for a dataset. Most commonly used texts are in the form of Question, Caption, Sentence with exceptions of a Lesson and a Paragraph in TQA and VLQA respectively. 
    \item \textbf{Task} represents the broad categorization defined by the NLP and Computer Vision community for each vision-language problem. Most tasks are in the form of question answering, popularly known as VQA. Additionally, if the task focuses on a particular reasoning skill needed to solve the dataset (e.g. counting, spatial reasoning, understanding text within images) or requires a domain specific knowledge, (charts, science, geometry, commonsense or world knowledge) is mentioned alongside.  
    \item \textbf{Task Types} indicates whether a task can be solved as a Classification, Text generation or a Ranking problem. Classification tasks are commonly formed as a Multiple Choice (MC) or N-class classification. Open Ended (OE) answers (as strings or numeric) and Captions are standard mechanisms to evaluate text generation style tasks. Vision-language task for ShapeWorld is the only one which employs Scoring mechanism to represent confidence level in range [0,1].  
    
\end{enumerate}

\subsection{Dataset Creation Pipeline}
\label{sec:pipeline}
\begin{figure}[h!]
  \center
  \includegraphics[width=0.5\textwidth]{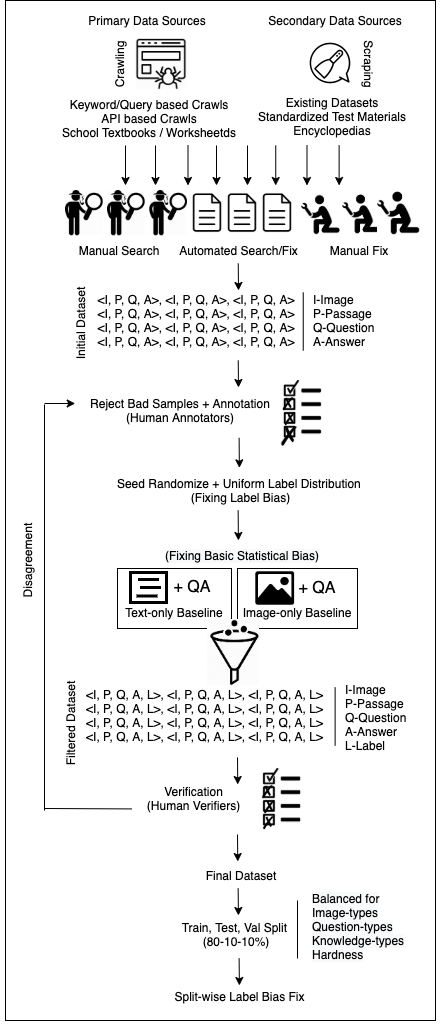}
  \caption{Data Collection, Processing and Integrity Steps implemented for construction of VLQA}
  \label{fig:flow}
\end{figure}

Figure \ref{fig:flow} illustrates the complete dataset creation pipeline. We divide overall process in 3 main stages- Data Collection, Annotation and Quality Control which is explained below;
\subsubsection{Data Collection and Post-processing}
VLQA task requires $<$Image, Passage, Question, AnswerChoices$>$ for each item in the corpus. To curate this dataset, we rely on data collection in two ways; One where variety of images are collected through crawling scripts that uses keyword search, existing APIs (flickr, twitter, newspapers, wikipedia, infographic websites etc.), images collected from documents and encyclopedias, which we refer to as primary data source. Then we manually find the relevant textual information in the context of the image and create questions based on it. We also tried generating templated images (like bar chart, pie chart, scatter plot etc.) from the tabular data obtained from CIA `world factbook' and WikiTables dataset. 
In the secondary data based method, we directly import items from human psychometric tests, exercises from school textbooks/handouts or existing vision-language datasets and then modify it in a way so that it fits the VLQA task. The data collection process included writing crawling/scraping scripts followed by combination of manual and automated search and fix such as,
\begin{itemize}
    \item replacing given textual/visual data with equivalent visual/textual counterparts respectively
    \item adding/removing partial information to/from text or visuals so that image and text do not contain identical information
    \item  creating factual or hypothetical situations around images 
\end{itemize}
 Then we standardize all collected information using above methods as multiple choice question-answers (MCQs) and get the initial version of the dataset. Our dataset includes all three kinds of visuals- Natural (everyday objects and scenes), Synthetic (artificially/program generated or templated figures) or Diagrams (imagery representing complex relationships between objects or phenomena). Each item in the VLQA dataset involves a considerable amount of text in passage, question and answer choices formed of diverse vocabulary of 33259 unique tokens. Also, these texts can involve facts, imaginary scenarios or their combination making it more realistic for real-world scenarios. This is how we compiled a large number of diverse items for the VLQA dataset in order to develop a  generic vision-language reasoning system. 
\subsubsection{Annotation} 
\begin{figure*}
  \includegraphics[width=\textwidth,height=9cm]{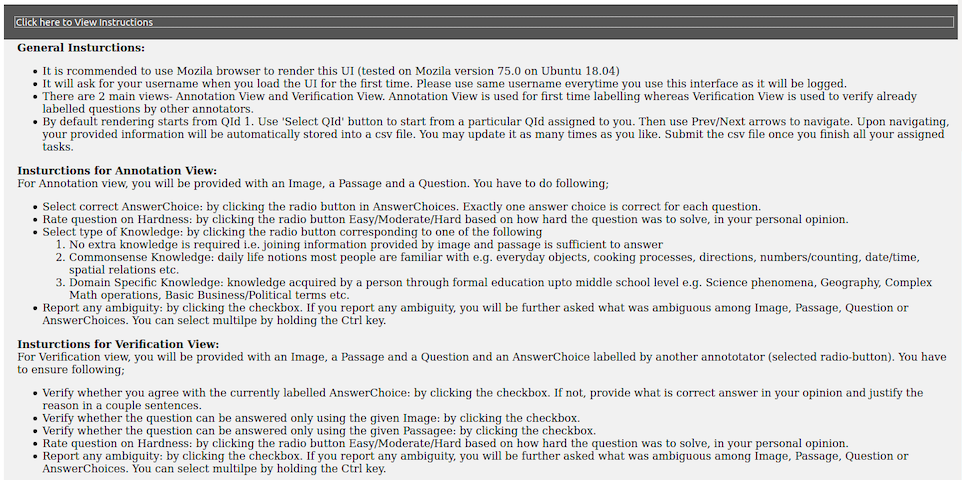}
  \caption{\textbf{3-fold instructions for annotators} Generic Instructions, Annotation Instructions and Verification Instructions.} 
  \label{fig:annot3}
\end{figure*}

\begin{figure*}
  \center
 \includegraphics[width=\textwidth]{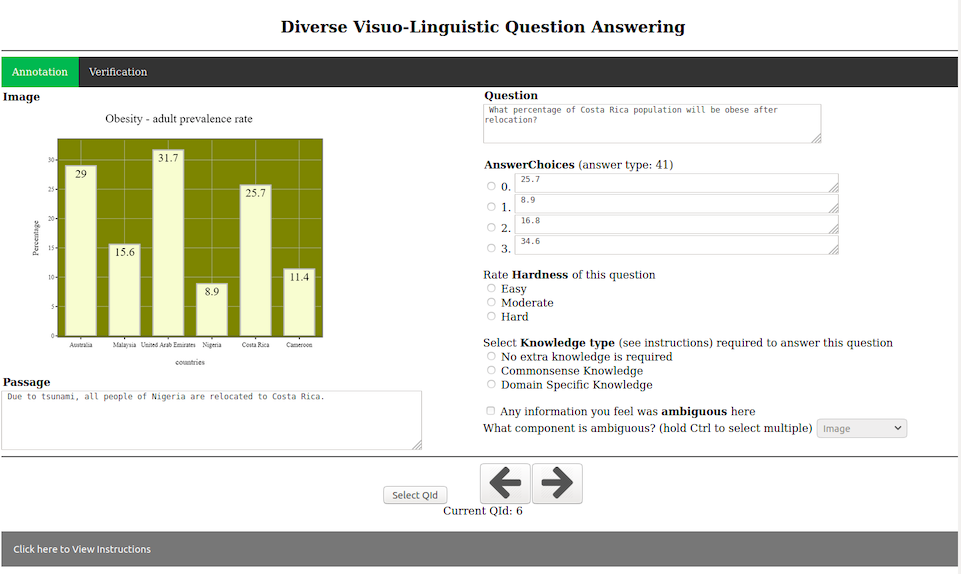}
  \caption{\textbf{Annotation View} is used for first time labelling of dataset items. $<$I, P, Q, A$>$ will be rendered in the UI after initial dataset formation. User has to determine the correct choice, categorize item based on knowledge type, rate for hardness and report ambiguity (if any).    
  }
  \label{fig:annot1}
\end{figure*}

All data items obtained from primary sources require  annotation as questions are created manually. For items obtained through secondary methods, annotation is required only if originally imported modalities were perturbed. Our crowd worker interface was designed using  Python-Flask\footnote{https://flask.palletsprojects.com/en/1.1.x/} and deployed on a local server. The data entered by annotators is then logged into Comma Separated Value (CSV) files in a structured format. Annotators were clearly instructed (as per figure \ref{fig:annot3}) about the annotation procedure and it was known to them that exactly one answer choice is correct for each item in our dataset. During first round of annotations, annotators were allowed to reject bad samples based on following two things; first, image and passage must not represent identical information and second, a question must not be answerable without looking at image and passage by marking it ambiguous as shown in Figure \ref{fig:annot1}. 

\begin{figure*}
  \center
  \includegraphics[width=\textwidth]{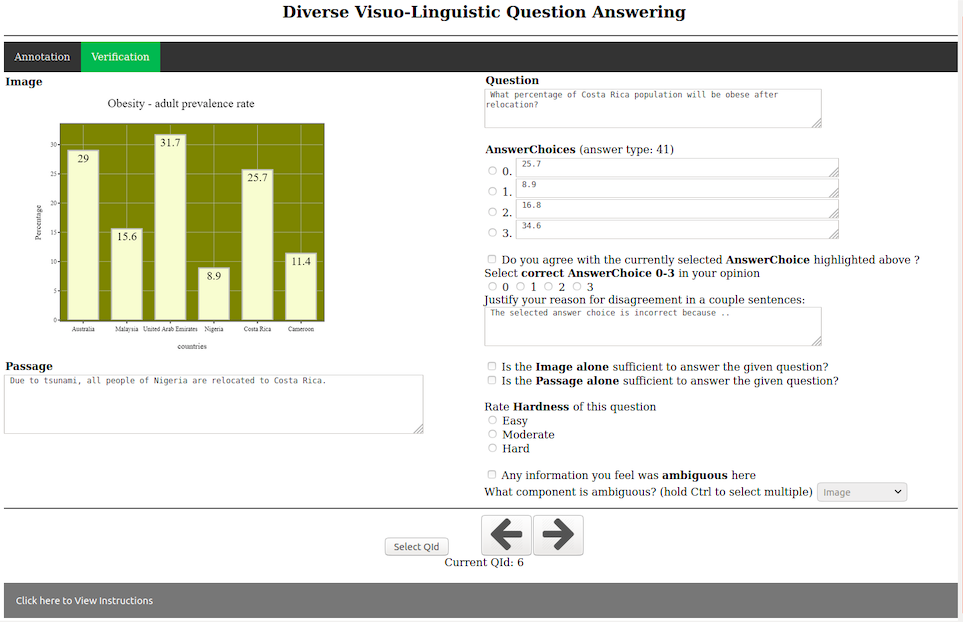}
  \caption{\textbf{Verification View} is used as a mechanism for inter-annotator agreement about the ground-truth label. $<$I, P, Q, A, L$>$ will be rendered in the UI post image-only and text-only baseline filtering. User checks for the correctness of label, rate for hardness and report ambiguity (if any).   
  }
  \label{fig:annot2} 
 
\end{figure*}

\subsubsection{Quality Control and Bias Mitigation}
Since we focus on the task of joint reasoning, we have to ensure that all our data items must use both image and passage. For the quality control purposes, we want to remove the samples which can be answered correctly by the state-of-the-art models in the absence of one of the modalities due to underlying bias it has learned from the training data. Therefore, we create 3 baselines- question-only (simply takes Q and predicts answer from choices A), passage-only (considers P as a context, takes Q and predicts answer from choices A) and image-only (considers I as a context, takes Q and predicts answer from choices A). We get predictions for whole data using these baselines. We repeat this experiment for 3 times by shuffling answer choices with a fixed seed. If a question can be answered correctly by any baseline in all trials, We remove such samples. Performance for these baselines is reported in Table \ref{tab:tab2}. The poor performance of these baselines indicate that the VLQA dataset requires models to jointly understand both image and text modalities.

Finally, we perform another round of manual quality check. We instruct workers to first try to answer a question just based on images and then try to answer a question based on only using text passage. If a question can be answered using a single modality, we suggest annotators to mark the checkbox as shown in Figure  \ref{fig:annot2}. Finally, we look over all bad samples and either provide a fix or remove, on a  case-by-case basis.

\begin{figure}
  \center
 \includegraphics[width=0.4\textwidth]{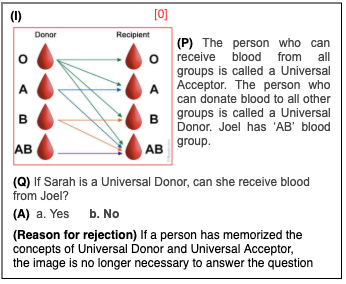} \\
 \includegraphics[width=0.4\textwidth]{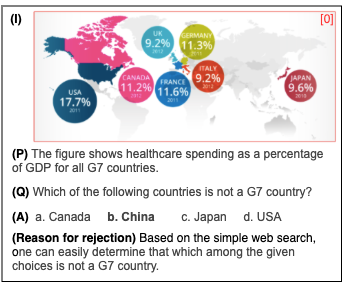}
  \caption{Example of 2 rejected VLQA samples with explanation for rejection.    
  }
  \label{fig:rej}
\end{figure}

We initially curated $\sim$12000 image-passage-qa pairs. During the annotation process, $\sim$700 were reported ambiguous, out of which we removed $\sim$500 and remaining $\sim$200 were modified and added back. By quality check process through baselines, we removed another $\sim$1900 samples. In the verification stage, we further removed $\sim$350 samples, and ended up with a dataset of 9267 samples eventually. Two rejected examples can be seen in Figure  \ref{fig:rej}, with explanation of reason for removal.

\clearpage
\onecolumn
\subsection{Format of Annotations provided for VLQA Dataset and explanation of each field
}
\begin{lstlisting}[language=json]
{
  "qid": 1,
  "images": [1.png,2.png,..],
  "multiple_images" : True/False,
  "passage": "This is a sample text passage.",
  "question": "Is this a sample question?",
  "answer_choices": ["choice0", "choice1", "choice2", "choice3"],
  "answer": 0/1/2/3,
  "image_type": "Natural"/"Templated"/"Freeform"
  "image_subtype": "Bar"/"Pie"/..,
  "answer_type": "4way_text",
  "multistep_inference": True/False,
  "reasoning_type": ["Deductive","Math"],
  "ext_knowledge": True/False,
  "ext_knowledge_type": "Commonsense"
  "ext_knowledge_text": "This is external knowledge required.",
  "ocrtokens": ["text","tokens","inside","image"],
  "image_source": "http://www.image/obtained/from/url/xyz",
  "passage_source": "wikipedia",
  "difficulty_level": "hard"/"easy"/"moderate",
  "split": "train"/"test"/"val"
}
\end{lstlisting}
\begin{itemize}
 
\item \textbf{qid}: Unique identifier for the item from 1 to 9267

\item \textbf{images}: Visual modality for the dataset item as a list of image file names, which will be assigned unique identifiers [0],[1],[2],.. and composed as a single file by merging (in order left to right) 

\item \textbf{multiple\_images}: Boolean field suggesting whether or not an item has multiple images

\item \textbf{passage}: Textual modality for the dataset item, typically consisting of 1-5 sentences. 

\item \textbf{question}: Question in natural language aiming to assess joint reasoning capability of a person/model

\item \textbf{answer\_choices}: Answer choices for a multiple choice question (MCQ) which can be short phrases, numeric, sentence, boolean or image (referred as a detection tag [0],[1],[2],..)

\item \textbf{answer}: Integer 0-3 corresponding to answer\_choices suggesting the ground-truth label for a question 

\item \textbf{image\_type}: Categorization of images based on whether they are ``Natural", ``Templated" (structured) or ``Freeform" (unstructured and not natural)

\item \textbf{image\_subtype}: "Templated" images are further classified in 20 subtypes listed as follows; \\
``Bar" (includes Simple/Stacked/Grouped), ``Pie" (or Donut chart), ``Scatter", ``Line", ``Area", ``Bubble" , ``Radar", ``VennDiagrams", ``Timelines", ``Hierarchies" (or Trees), ``Maps", ``Tables" (or Matrix), ``Cycles", ``Processes", ``Heatmaps", ``DirectedGraphs", ``UndirectedGraphs", ``FlowCharts", ``SankeyDiagram", ``CoordinateSystems" (this field will be empty for ``Natural" and ``Freeform" images)

\item \textbf{answer\_type}: Classification of item based on 5 answer types listed as follows;
\begin{enumerate}

\item  \textbf{4-way text (4wT)}: [``text0",``text1",``text2",``text3"] \\
   One need to select the correct alpha-numeric choice among 4 choices based on the scenario described in question, passage and image
\item  \textbf{4-way Sequencing (4wS)}: [``I-II-IV-III",``I-IV-III-II",``II-III-I-IV",``II-I-IV-III"]   \\
   Consider 4 steps (I-IV) in a process which is represented as a  combination of image and text, and jumbled up. One has to select the correct order of events from given choices. 
\item  \textbf{4-way image (4wI)}: [``[1]",``[2]",``[0]",``[3]"] where [x] are image detection tags \\
   One needs to select the correct image among 4 choices based on the scenario described in question and passage. Images are referred through detection tags [0],[1],[2],[3]. 
\item  \textbf{2-way image (2wI)}: [``[1]",``[0]"] where [x] are image detection tags \\
   One needs to select the correct image among 2 choices based on the scenario described in question and passage. Images are referred through detection tags [0],[1]. 
\item  \textbf{Binary Classification (Bin)}: [``True",``False"] or [``No",``Yes"] \\
   One needs to determine whether or not the text in question is true or false with respect to the given visuo-linguistic context.
 
\end{enumerate}

\item \textbf{multistep\_inference}: Boolean field suggesting whether or not the question requires multiple inference steps to correctly answer the question

\item \textbf{reasoning\_type}: A list of reasoning skills required to solve given question, most frequently observed types are listed as follows; 
\begin{enumerate}
    \item ``InfoLookup" (look for a specific information or conditional retrieval)
\item ``Temporal" (reasoning with respect to time)
\item ``Spatial" (reasoning with respect to space) 
\item ``Deductive" (given a generic principle, deduce a conclusion for a specific case and vice-versa) 
\item ``Abductive" (finding most plausible explanation with respect to given set of observations) 
\item ``Mathematical" (arithmetic, trends, minimum/maximum, counting, comparison, complement, fractions, percentages etc.) 
\item ``Logical" (conjunction, disjunction, logical negation, existential quantifiers etc.) 
\item ``Causality" (cause-effect relationship) 
\item ``Analogy" (comparison for the purpose of explanation or clarification, different from numerical comparison)
\item ``Verbal" (synonym/antonym, subclass/superclass, vocabulary, verbal negation etc.) 
\end{enumerate}

\item \textbf{ext\_knowledge}: Boolean field suggesting whether or not the question requires any external knowledge beyond what is provided in visuo-linguistic context

\item \textbf{ext\_knowledge\_type}: Classification of required external knowledge as follows; 
\begin{enumerate}
    \item Commonsense: Facts about the everyday world, which most people know.
    \item Domain Specific knowledge: Knowledge acquired through formal study (we limit our domain specific knowledge to middle-school level)
\end{enumerate}

\item \textbf{ext\_knowledge\_text}: Manually written justification of required external knowledge

\item \textbf{ocrtokens}: List of OCR extracted tokens from images and manually corrected if erroneous, just in case some systems would like incorporate OCR based features

\item \textbf{image\_source}: Source/Weblink from which image is retrieved (original image might be altered in some cases before using it for this dataset)

\item \textbf{passage\_source}: Source/Weblink of passage (if retrieved from some source), empty if passage written manually

\item \textbf{difficulty\_level}: Difficulty level of question, decided by majority annotator opinion classified as "Hard", "Easy" or "Moderate"

\item \textbf{split}: "Train", "Test" or "Val" partition, whichever the sample belongs to 

\end{itemize}

\begin{figure*}
\subsection{Additional Dataset Samples}
\label{sec:moreex}
We provide more examples from the VLQA dataset to visualize the diversity offered by the corpus and importance of joint reasoning to derive conclusions for real-world scenarios. 
 \center
  \includegraphics[width=\textwidth]{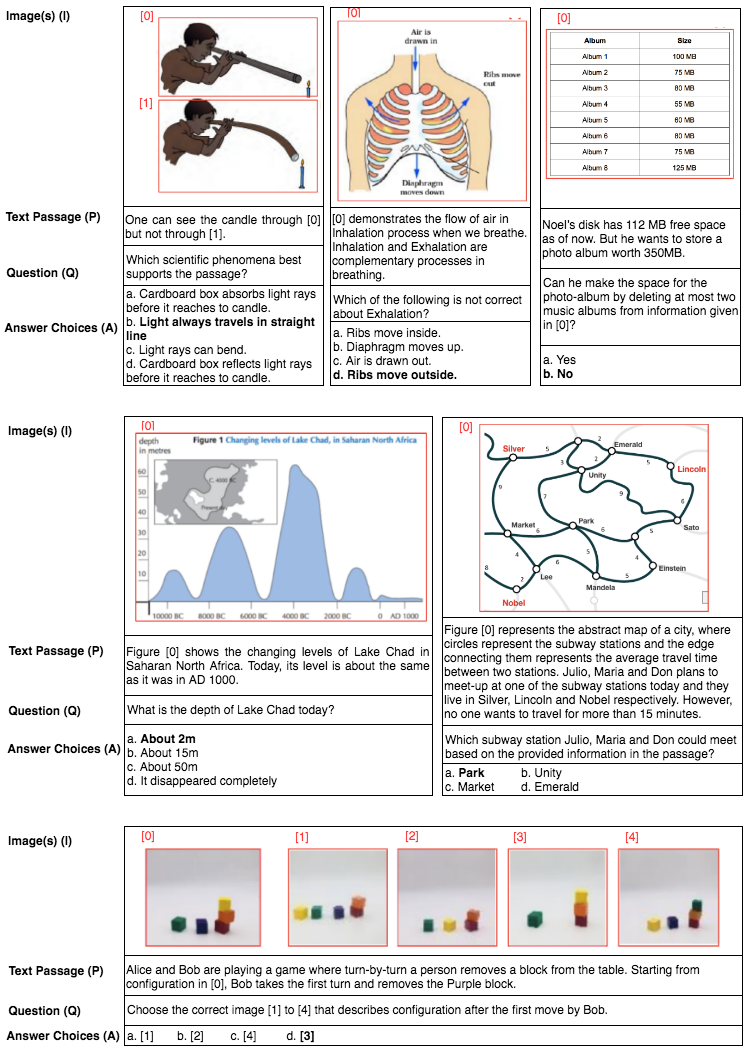}
  \label{fig:app1}
\end{figure*}

\begin{figure*}
\textbf{Additional Dataset Samples- Continued}
  \center
  \includegraphics[width=\textwidth]{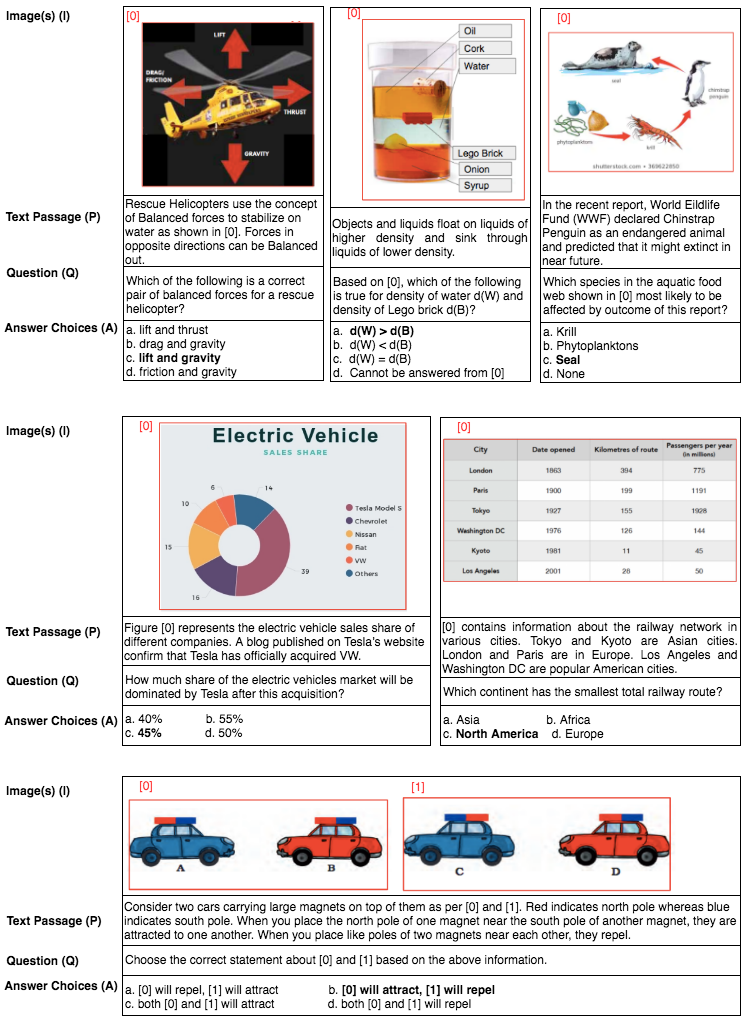}
  \label{fig:app2}
\end{figure*}

\twocolumn

\section{Supplemental Material}
\label{sec:supplemental}

\paragraph{Computing Infrastructure}
All experiments are done over Tesla V100-PCIE-16GB GPU. 

\subsection{Converting Visual COPA dataset into Image-Image Entailment Task}
VCOPA dataset contains visual questions with three images- one labelled as premise (P) image, and other two as alternatives (H1 \& H2). The task is to identify plausible alternative image related to the premise. We convert VCOPA item into Image-Image entailment task as 2-way classification as below; \\
Given VCOPA sample: \\ $<$P, H1, H2$>$ $|$ label: H1 (i.e. plausible choice) \\
Converted Image-Image Entailment samples: \\
$<$P, H1$>$ $|$ label: Entailment \\
$<$P, H2$>$ $|$ label: Contradiction \\
Then a custom 3-layer network is trained to maximize the above classification  

\subsection{Converting Visual COPA dataset into Text-Image Entailment Task}
Similar to above, we convert VCOPA item into Text-Image entailment with additional Image Captioning module. \\
Given VCOPA sample: \\ $<$P, H1, H2$>$ $|$ label: H1 (i.e. plausible choice) \\
Using the Image Captioning module, get a caption for P i.e. C\_P, while keeping H1 and H2 in the image format itself. 
Converted Text-Image Entailment samples: \\
$<$C\_P, H1$>$ $|$ label: Entailment \\
$<$C\_P, H2$>$ $|$ label: Contradiction \\
Then a custom 3-layer network is trained to maximize the above classification.  

\subsection{Model Parameters}
Detailed summary of various components implemented for this paper - Brief description, Reference Code Link and Parameters provided in Table \ref{tab:tab6}.

\begin{landscape}
\begin{table}
\begin{tabular}{@{}lllll@{}}
\toprule
\multicolumn{2}{l}{}                                  & \textbf{Usage}                                                                                                                                                                                                       & \textbf{Ref.}                                                                                                          & \textbf{Setting}                                                                                                                                                                \\ \midrule
\multicolumn{2}{l}{\textbf{Quality Check}} &                                                                                                                                                                                                                      &                                                                                                                        &                                                                                                                                                                                 \\
                 & Q-only: RoBERTA                    & RoBERTa large + RACE Ft. + ARC Ft. - Predict on VLQA $<$Q,A$>$                                                                                                                                                          & \href{https://leaderboard.allenai.org/arc/submission/blcotvl7rrltlue6bsv0}{Link}                                                    & \begin{tabular}[c]{@{}l@{}}RoBERTa Large ft. on RACE with \\ LR=1e-5, BS=16, WD=0.1, LRD=Linear, EP=4, WR=0.06\\ Further ft. on ARC with BS = 8, EP = 4, LR = 1e-5\end{tabular} \\
                 & P-only: ALBERT                     & ALBERT-xxl + RACE Ft. - Predict on VLQA $<$P,Q,A$>$                                                                                                                                                                     & \href{https://github.com/google-research/albert}{Link}                                                                              & ALBERT-xxl (v2) ft. on RACE with LR=1e-5, BS=32, DR=0                                                                                                                           \\
                 & I-only: LXMERT                 & LXMERT + VQA Ft. - Predict on VLQA $<$I,Q,A$>$                                                                                                                                                                          & \href{https://github.com/airsplay/lxmert}{Link}                                                                                    & LXMERT ft. on VQA with BS=32, LR=5e-5, EP=4                                                                                                                                     \\ \midrule
\multicolumn{2}{l}{\textbf{Pre-trained VL}} &                                                                                                                                                                                                                      &                                                                                                                        &                                                                                                                                                                                 \\
                 & VL-BERT                            & VL-BERT + VQA Ft. + VLQA Ft. $<$I,P+Q,A$>$                                                                                                                                                                               & \href{https://github.com/jackroos/VL-BERT}{Link}                                                                                    & \begin{tabular}[c]{@{}l@{}}VL-BERT ft. on VQA with EP=20, BS=256, LR=1e-4, WD=1e-4\\ Ft. on VLQA with BS=16, LR=2e-5, EP=15\end{tabular}                                        \\
                 & VisualBERT                         & VisualBERT + VQA Ft. + VLQA Ft. $<$I,P+Q,A$>$                                                                                                                                                                            & \href{https://github.com/uclanlp/visualbert}{Link}                                                                                  & \begin{tabular}[c]{@{}l@{}}VisualBERT ft. on VQA with BS=32, LR=2e-5, EP=10\\ Ft. on VLQA with BS=16, LR=1e-5, EP=10\end{tabular}                                               \\
                 & ViLBERT                            & ViLBERT + VQA Ft. + VLQA Ft. $<$I,P+Q,A$>$                                                                                                                                                                               & \href{https://github.com/facebookresearch/vilbert-multi-task}{Link}                                                                 & \begin{tabular}[c]{@{}l@{}}ViLBERT ft. on VQA with BS=32, LR=1e-5, EP=20, WR=0.1\\ Ft. on VLQA with BS=32, LR=1e-5, EP=10\end{tabular}                                          \\
                 & LXMERT                             & LXMERT + VQA Ft. + VLQA Ft. $<$I,P+Q,A$>$                                                                                                                                                                                & \href{https://github.com/airsplay/lxmert}{Link}                                                                                     & \begin{tabular}[c]{@{}l@{}}LXMERT ft. on VQA with BS=32, LR=5e-5, EP=4\\ Ft. on VLQA with BS=16, LR=5e-5, EP=8\end{tabular}                                                     \\
               \\ \midrule
\multicolumn{2}{l}{\textbf{Proposed HOLE}}   &                                                                                                                                                                                                                      &                                                                                                                        &                                                                                                                                                                                 \\
                 & Text Entailment                    & RoBERTa large + MNLI Ft.                                                                                                                                                                                             & \href{https://github.com/pytor ch/fairseq/tree/master/examples/roberta}{Link}                                                       & \begin{tabular}[c]{@{}l@{}}RoBERTa Large ft. on MNLI with \\ LR=1e-5, BS=16, WD=0.1, LRD=Linear, EP=10, WR=0.06\end{tabular}                                                    \\
                 & I-T Entailment                     & Bilateral Multi-Perspective Matching (BiMPM) on SNLI                                                                                                                                                                 & \href{https://github.com/claudiogreco/coling18-gte}{Link}                                                                           & --                                                                                                                                                                              \\
                 & I-I Entailment                     & \begin{tabular}[c]{@{}l@{}}VCOPA dataset converted into Image-Image Entailment \\ Trained custom 3-layer network for 2-class classification\end{tabular}                                                             & \href{https://github.com/antest1/VCOPA-Dataset}{Link}                                                                               & LR=1e-5, BS=16, OP=Adam, WD=0.1, EP=10     
                 
                                                          \\
                 & T-I Entailment                     & \begin{tabular}[c]{@{}l@{}}VCOPA dataset converted into Text-Image Entailment + Captioning \\ Trained custom 3-layer network for 2-class classification \end{tabular}                                                             & \href{https://github.com/antest1/VCOPA-Dataset}{Link}                                                                               & LR=1e-5, BS=16, OP=Adam, WD=0.1, EP=10     
                 
                 \\
                 & LXMERT                             & LXMERT + VQA Ft. + VLQA Ft.                                                                                                                                                                                          & \href{https://github.com/airsplay/lxmert}{Link}                                                                                     & LXMERT ft. on VQA with BS=32, LR=5e-5, EP=4                                                                                                                                     \\
                 & ALBERT+LXMERT                      & \begin{tabular}[c]{@{}l@{}}ALBERT-xxl + RACE Ft. - Predict on VLQA $<$P,Q$>$ to get A'\\ Generate Q' as "Where is A'?" and substitute A' with above string\\ LXMERT + VQA Ft.. - Predict on VLQA $<$I,Q',A>\end{tabular} & \begin{tabular}[c]{@{}l@{}}\href{https://github.com/google-research/albert}{Link}\\ \href{https://github.com/airsplay/lxmert}{Link}\end{tabular} & \begin{tabular}[c]{@{}l@{}}ALBERT-xxl (v2) ft. on RACE with LR=1e-5, BS=32, DR=0\\ LXMERT ft. on VQA with BS=32, LR=5e-5, EP=4\end{tabular}                                                                                                                    \\    \bottomrule
\end{tabular}
\caption{\label{tab:tab6} Links to the reference code used for the paper and relevant parameters \\ BS - Batch Size, DR - Dropout Rate, EP - Epochs, LR - Learning Rate, LRD - Learning Rate Decay, WD - Weight Decay, WR  - Warmup Ratio, Ft. - Manual Finetuning
}
\end{table}
\end{landscape}

\end{document}